\documentclass[conference]{IEEEtran}
\IEEEoverridecommandlockouts
\usepackage{cite}
\usepackage{amsmath,amssymb,amsfonts}
\usepackage{algorithmic}
\usepackage{graphicx}
\usepackage{textcomp}
\usepackage{xcolor}
\usepackage{multirow}
\usepackage{xurl}
\usepackage{float}
\usepackage{booktabs}

\def\BibTeX{{\rm B\kern-.05em{\sc i\kern-.025em b}\kern-.08em
    T\kern-.1667em\lower.7ex\hbox{E}\kern-.125emX}}
\begin{document}

\title{Unveiling the Heart-Brain Connection: An Analysis of ECG in Cognitive Performance}

\author{
\IEEEauthorblockN{1\textsuperscript{st} Akshay Sasi}
\IEEEauthorblockA{
\textit{Digital University Kerala} \\
Thiruvananthapuram, India \\
\texttt{akshaysasi12.knr@gmail.com} \\
ORCID: 0009-0009-3708-554X
}
\and
\IEEEauthorblockN{2\textsuperscript{nd} Malavika Pradeep}
\IEEEauthorblockA{
\textit{Digital University Kerala} \\
Thiruvananthapuram, India \\
\texttt{malavikapradeep2001@gmail.com} \\
ORCID: 0009-0005-9215-3542
}
\and
\IEEEauthorblockN{3\textsuperscript{rd} Nusaibah Farrukh}
\IEEEauthorblockA{
\textit{Digital University Kerala} \\
Thiruvananthapuram, India \\
\texttt{nusaibah.farrukh@gmail.com} \\
ORCID: 0009-0004-1237-2693
}
\and
\IEEEauthorblockN{4\textsuperscript{th} Rahul Venugopal}
\IEEEauthorblockA{
\textit{Centre for Consciousness Studies, NIMHANS} \\
Bangalore, India \\
\texttt{rhlvenugopal@gmail.com} \\
ORCID: 0000-0001-5348-8845
}
\and
\IEEEauthorblockN{5\textsuperscript{th} Elizabeth Sherly}
\IEEEauthorblockA{
\textit{Digital University Kerala} \\
Thiruvananthapuram, India \\
\texttt{sherly@duk.ac.in} \\
ORCID: 0000-0001-6508-950X
}
}

\maketitle

\begin{abstract}
Understanding the interaction of neural and cardiac systems during cognitive activity is critical to advancing physiological computing. Although EEG has been the gold standard for assessing mental workload, its limited portability restricts its real-world use. Widely available ECG through wearable devices proposes a pragmatic alternative. This research investigates whether ECG signals can reliably reflect cognitive load and serve as proxies for EEG-based indicators. In this work, we present multimodal data acquired from two different paradigms involving working-memory and passive-listening tasks. For each modality, we extracted ECG time-domain HRV metrics and Catch22 descriptors against EEG spectral and Catch22 features, respectively. We propose a cross-modal XGBoost framework to project the ECG features onto EEG-representative cognitive spaces, thereby allowing workload inferences using only ECG. Our results show that ECG-derived projections expressively capture variation in cognitive states and provide good support for accurate classification. Our findings underpin ECG as an interpretable, real-time, wearable solution for everyday cognitive monitoring.
\end{abstract}


\section{Introduction}
\begin{figure}[htbp]
    \centering
    \includegraphics[width=1\linewidth]{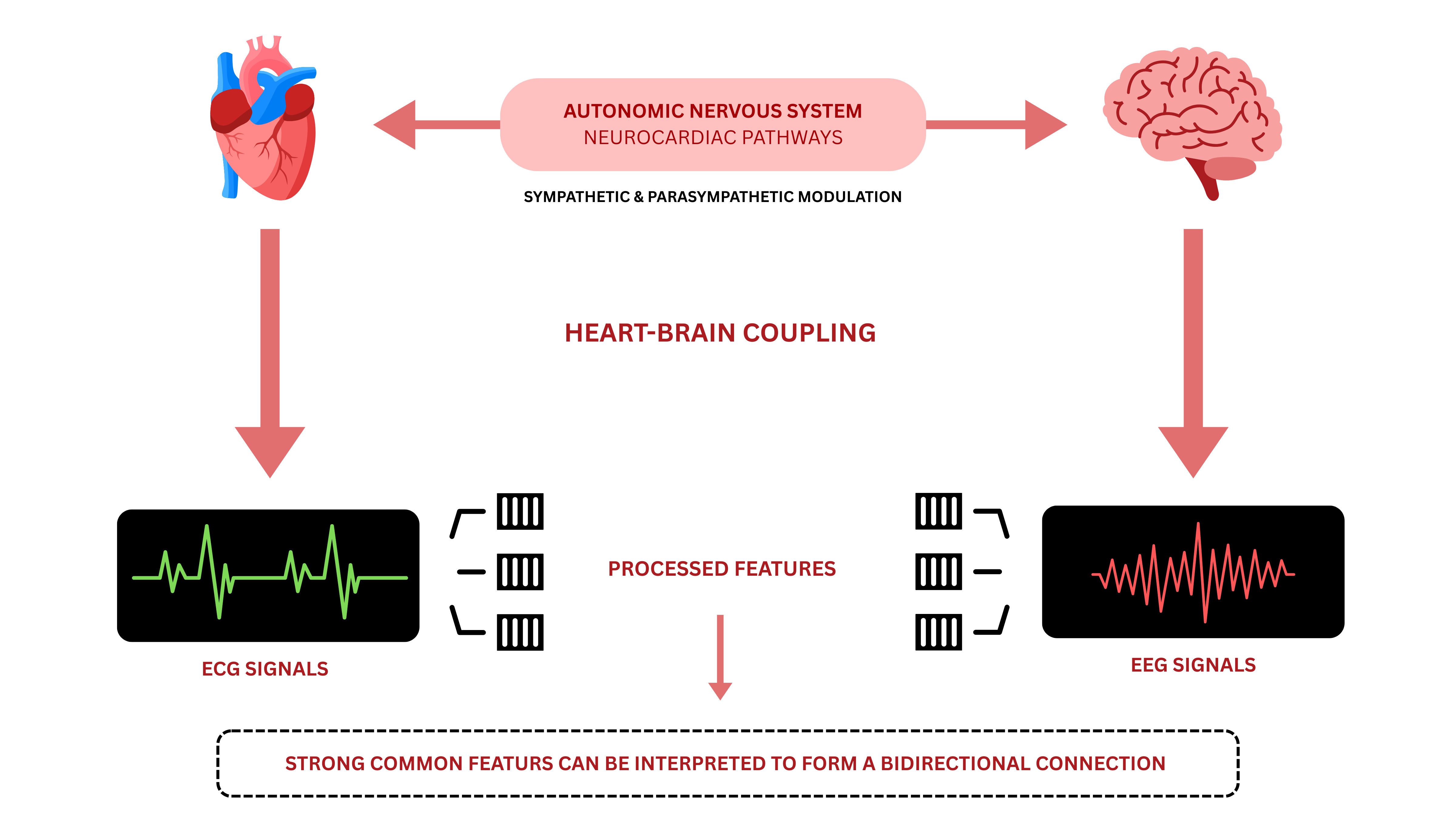} 
    \caption{Heart-Brain Coupling}
    \label{fig:heart_brain_coupling}
\end{figure}

The dynamic, bidirectional communication between the heart and the brain, mediated by the autonomic nervous system, impacts cognitive and emotional states significantly \cite{hrv_brain_heart}. A fundamental noninvasive marker of this interaction is Heart Rate Variability (HRV), which reflects a person's ability to adapt to various physiological and mental demands. For illustration of this bidirectional coupling see Fig.~\ref{fig:heart_brain_coupling}.

Cognitive load assessment plays an important role in a wide variety of areas, from aviation to education. Though EEG is considered the gold standard for cortical activity measurement, its inconvenient hardware and susceptibility to various types of artifacts make it unsuitable for continuous monitoring in daily life \cite{2017eeg,eeg2023lstm}. ECG-based HRV represents a promising alternative, with ECG sensors now integrated into many consumer wearables and associated with neurocognitive and emotional regulation \cite{Brainheart,2024cardiaclink,hestad2022heart}.

This challenge calls for the investigation of robust physiological analysis. Standardized HRV features are typically only statistically stable for recordings of a long duration,  thus ill-suited for the small observation windows ($<$5s) needed for real-time cognitive load detection. For this purpose,  we employ the Catch22 feature set: a handpicked selection of 22 robust time series characteristics that captures complex, non-linear dynamics detectable even in ultra-short intervals.

This paper offers three major contributions. First, we empirically show that Catch22 features have higher discriminative power compared with conventional HRV metrics. Second, we validate a novel fusion classifier based on transfer learning which integrates ECG and EEG modalities. Third, through the cross-modal framework, we provide evidence for the heart-brain functional coupling, confirming that ECG-derived signals can serve as a reliable proxy for EEG in estimating cognitive workload during an auditory Digit Span Task (OpenNeuro ds003838).

\subsection{Dataset and Study Design}
The OpenNeuro data contains synchronized ECG and EEG recordings from 65 healthy adults in four Digit Span Task conditions: Just Listen (baseline),  Five-digit (low load),  Nine-digit (medium load),  and Thirteen-digit (high load). Retention intervals were extracted and structured into a 5D array to enable synchronized cross-modal analysis. Cognitive overload is of critical concern in fields like health care and aviation, so utilizing the wearables-accessible nature of ECG enables a non-invasive path toward continuous cognitive health tracking,  bridging a gap between clinical-grade neural monitoring and daily-life feasibility.

\section{Related Work}
Recent efforts in the field of heart–brain interaction have increasingly concentrated on cross-modal computational models using heart rate variability from ECG in conjunction with EEG in order to estimate the cognitive state. Multimodal fusion frameworks have indeed demonstrated that peripheral cardiac dynamics can reflect neural activity, especially in physiological stress and workload estimation \cite{cogML,DL1,wang2023framework,DL2}.

These include a variety of fusion approaches, using machine learning and deep models, such as CNN-based and XGBoost pipelines, which have shown very strong classification performance on both memory and attention tasks within cognitive load research. These collectively suggest that multimodal learning is more robust compared to unimodal HRV methods, which remain sensitive to noise and task dependent.

Beyond workload estimation,  transfer learning and cross-domain mapping have been popular approaches for bridging ECG and EEG representations \cite{transferlearning}.

Recent models of EEG→ECG and cross-physiology report successful generalization for movement intention and saccade-based inhibitory control, while hybrid physiological deep models \cite{hybrid_future,2021deepTF} and cross-domain transfer studies \cite{transferlearning} further support scalability of multimodal inferences.
Although progress has been made, challenges persist due to data scarcity, modality mismatch, and limited real-time evaluability, hence motivating our proposed ECG→EEG transfer learning approach for cognitive monitoring.

\section{Methodology}
We used the OpenNeuro dataset ds003838 \cite{digitspan}, which includes synchronized EEG and ECG recordings of an auditory digit span task in 86 participants. After data quality assessment, we included 65 participants for analysis who had both clean EEG and ECG signals.

\subsection{Data Recording Protocol}
\begin{figure}[htbp]
    \centering
    \includegraphics[width=0.95\linewidth]{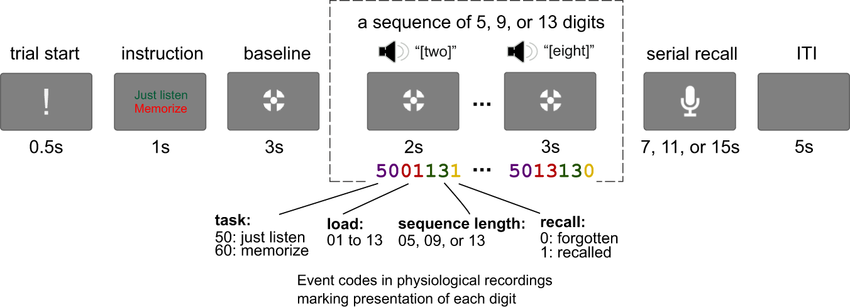} 
    \caption{Digit span task experimental paradigm.}
    \label{fig:digitspan}
\end{figure}

Participants performed digit-span trials (5,  9,  13 digits + baseline Just Listen) with baseline,  encoding,  and 3s retention phases,  with only retention used for analysis. A visual representation of the task is provided in Figure \ref{fig:digitspan}.

\subsection{Working Model}
\begin{figure}[htbp]
    \centering
    \includegraphics[width=0.95\linewidth]{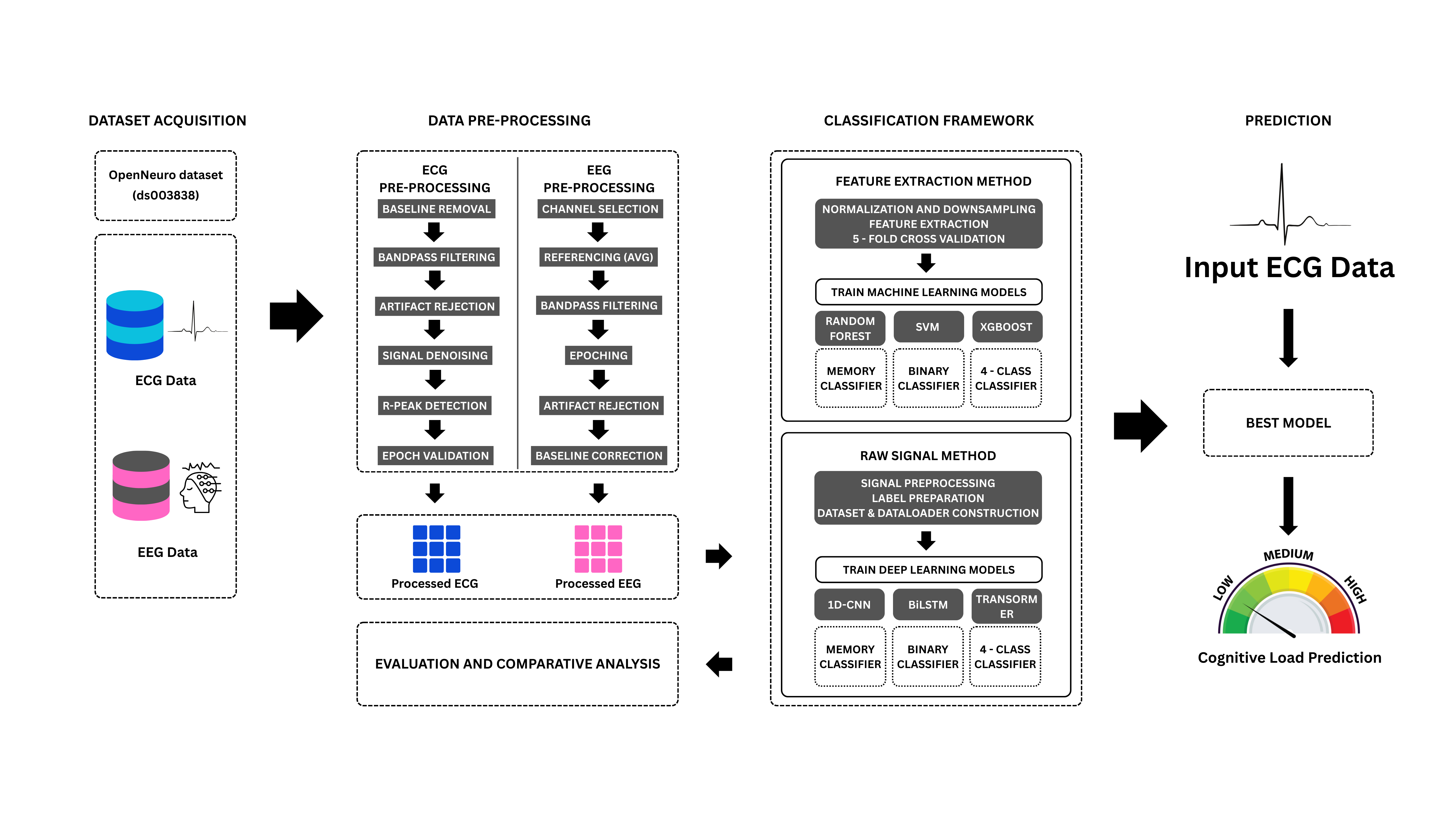} 
    \caption{Architecture Diagram of the proposed framework.}
    \label{fig:architecture}
\end{figure}

Our end-to-end framework (Fig. \ref{fig:architecture}) processes synchronized ECG and EEG signals from the digit span task to predict cognitive load. The pipeline begins with modality-specific preprocessing. We then pursue two parallel classification approaches: (1) Traditional ML models (Random Forest,  SVM,  XGBoost) trained on hand-crafted features,  and (2) Deep Learning models (1D-CNN,  BiLSTM,  Transformer) trained on raw signals.

\subsection{ECG Pre-processing}
Raw ECG signals were preprocessed using a second-order bandpass filter (0.5--40 Hz) to remove baseline drift and high-frequency noise. The filtering operation follows the convolution process defined in Eq.~\ref{eq:convolution}.
\begin{equation}
y(t) = x(t) * h(t)
\label{eq:convolution}
\end{equation}

We utilized the NeuroKit2 \cite{neurokit2} library for robust signal processing. R-peak detection was performed using adaptive thresholding (${nk.ecg\_peaks()}$). RR intervals were computed as the time difference between successive R-peaks ($RR_i = R_{i+1} - R_i$). Artifact correction was applied to the RR series using Median Absolute Deviation (MAD) filtering to remove outliers.

From the cleaned intervals,  time-domain Heart Rate Variability (HRV) features were extracted,  including SDNN and RMSSD:
\\
\begin{equation}
SDNN = \sqrt{\frac{1}{N-1} \sum_{i=1}^{N}(RR_i - \bar{RR})^2}
\label{eq:sdnn}
\end{equation}

\begin{equation}
RMSSD = \sqrt{\frac{1}{N-1} \sum_{i=1}^{N-1}(RR_{i+1} - RR_i)^2}
\label{eq:rmssd}
\end{equation}
\\

HRV metrics (SDNN,  RMSSD,  SD1,  SD2) and frequency-domain measures (LF,  HF,  LF/HF) were derived from RR intervals. Finally, processed data were structured into a 5D array (subject × condition × subcondition × trial × time).

\subsection{EEG Pre-processing}
EEG data were processed using MNE-Python \cite{GRAMFORT2014446}. The pipeline,  prioritizing robust artifact removal \cite{chaddad2023electroencephalography},  included:
\begin{itemize}
    \item Bandpass filtering (1--45 Hz) to retain cognitive-related bands (theta,  alpha,  beta).
    \item A 50 Hz Notch filter to remove line noise.
    \item Independent Component Analysis (ICA) to identify and remove ocular and muscle artifacts,  validated against EOG channels.
\end{itemize}
Cleaned signals were epoched to the retention window,  and memory-relevant channels selected: \texttt{['Fz',  'Pz',  'Cz',  'P3',  'P4']}. These epochs were then downsampled to 128 Hz,  baseline-corrected,  and stored in a 5D array for synchronized analysis.

\subsection{Feature Extraction}
We compared two distinct ECG feature extraction pipelines.

\subsubsection{HRV-Based Feature Extraction:}
HRV features were derived from the RR-interval series obtained after NeuroKit2-based R-peak detection. RR intervals were calculated in milliseconds using Equation \ref{eq:RR_interval}:
\begin{equation}
RR_i = \frac{(R_{i+1} - R_i) \times 1000}{f_s}
\label{eq:RR_interval}
\end{equation}
where $f_s$ is the sampling rate (250 Hz). Five key features were computed per trial: \textbf{MeanNN},  \textbf{SDNN},  \textbf{RMSSD},  \textbf{SD1},  and \textbf{SD2}. The non-linear features SD1 and SD2 were calculated as shown in Equations \ref{eq:sd1} and \ref{eq:sd2}:
\begin{equation}
SD1 = \sqrt{0.5} \times RMSSD
\label{eq:sd1}
\end{equation}
\begin{equation}
SD2 = \sqrt{2 \times SDNN^2 - 0.5 \times RMSSD^2}
\label{eq:sd2}
\end{equation}

In total,  27 features were extracted per trial from the ECG signal (5 HRV metrics + 22 Catch22 descriptors).

\subsubsection{Catch22 Feature Extraction:}
We also extracted the 'catch22' feature set \cite{article,c22_cardiac,c22ppg},  a standardized collection of 22 time-series characteristics spanning statistical,  dynamical,  and entropy-based domains. This was computed from the raw ECG signals per trial,  yielding a 22-dimensional feature vector:

\[
\textbf{f}_{\text{Catch22}} = [f_1,  f_2,  \dots,  f_{22}]
\]

Catch22 has proven effective on short physiological windows and wearable cardiac signals \cite{c22_cardiac}. Classical HRV-based approaches have also been used for online workload detection, providing a baseline for our time-series comparison \cite{hrv}.

\subsection{Classification Framework}
Models were trained for three tasks: (i) Memory Load (5/9/13),  (ii) Binary Load (baseline vs memory),  and (iii) Four-class classification. Feature-based ML (RF/SVM/XGBoost) used HRV and Catch22 features,  while 1D-CNN,  BiLSTM and Transformer networks were trained end-to-end on raw signals. For cross-modal fusion,  Catch22 was extracted from ECG and from each EEG channel (5 $\times$ 22 = 110-dim vectors),  enabling transfer experiments in both directions (ECG$\rightarrow$EEG and EEG$\rightarrow$ECG) \cite{yang2023cross,transferlearning,2021deepTF}.

\subsection{Model Evaluation Metrics}
Accuracy,  macro-F1 and weighted-F1 were used for evaluation,  along with ROC-AUC for binary tasks. Confusion matrices enabled inter-class error analysis. Comparative analyses were conducted across modalities,  features,  and model types.

\section{Results}
We evaluated our framework using ECG and EEG signals from the OpenNeuro dataset (ds003838) across three tasks: Memory Classifier (MC: Five,  Nine,  Thirteen),  Binary Classifier (BC: Just Listen vs. Memory),  and Four-Class Classifier (FC: Just Listen,  Five,  Nine,  Thirteen).

\subsection{ECG Classification with Classical ML}
We first compared models trained on 5 HRV features versus 22 Catch22 features (Table \ref{tab:ecg_classical}). Catch22 features dramatically outperformed traditional HRV features across all models (e.g.,  98.68\% vs 79.12\% in MC). Performance on the BC task was poor for both feature sets (59.78\% max),  suggesting difficulty distinguishing baseline "Just Listen" from memory states.

\begin{table}[htbp]
\caption{ECG Classification Performance (Classical ML Models)}
\label{tab:ecg_classical}
\begin{center}
\resizebox{\columnwidth}{!}{
\begin{tabular}{|c|c|c|c|c|}
\hline
\textbf{Task} & \textbf{Feature Set} & \textbf{Model} &
\textbf{Accuracy (\%)} & \textbf{Macro-F1} \\ \hline

\multirow{3}{*}{MC} & \multirow{3}{*}{HRV}
& RF      & 80.89 & 0.80 \\ \cline{3-5}
& & SVM     & 68.76 & 0.67 \\ \cline{3-5}
& & XGBoost & 79.12 & 0.79 \\ \hline

\multirow{3}{*}{MC} & \multirow{3}{*}{Catch22}
& RF      & 97.71 & 0.98 \\ \cline{3-5}
& & SVM     & 89.19 & 0.89 \\ \cline{3-5}
& & XGBoost & 98.68 & 0.99 \\ \hline

\multirow{3}{*}{BC} & \multirow{3}{*}{HRV}
& RF      & 50.33 & 0.48 \\ \cline{3-5}
& & SVM     & 52.52 & 0.49 \\ \cline{3-5}
& & XGBoost & 55.64 & 0.50 \\ \hline

\multirow{3}{*}{BC} & \multirow{3}{*}{Catch22}
& RF      & 58.38 & 0.53 \\ \cline{3-5}
& & SVM     & 44.55 & 0.44 \\ \cline{3-5}
& & XGBoost & 59.78 & 0.53 \\ \hline
\end{tabular}
}
\end{center}
\end{table}

\begin{figure}[htbp]
    \centering
    \includegraphics[width=0.95\linewidth]{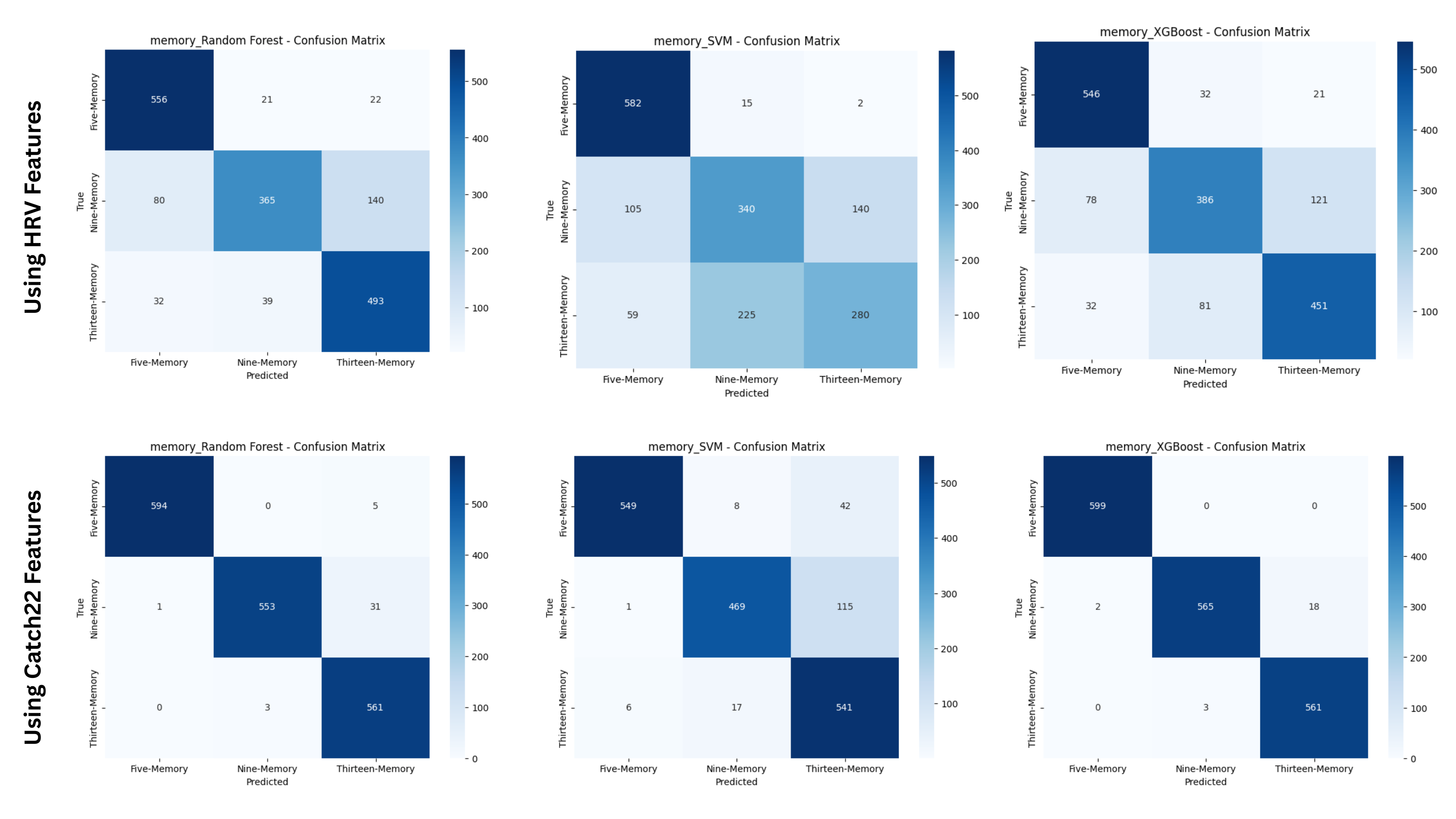} 
    \caption{Confusion Matrices for ECG Memory Classifier (MC).}
    \label{fig:C22vsHRV_memory}
\end{figure}

\begin{figure}[htbp]
    \centering
    \includegraphics[width=0.95\linewidth]{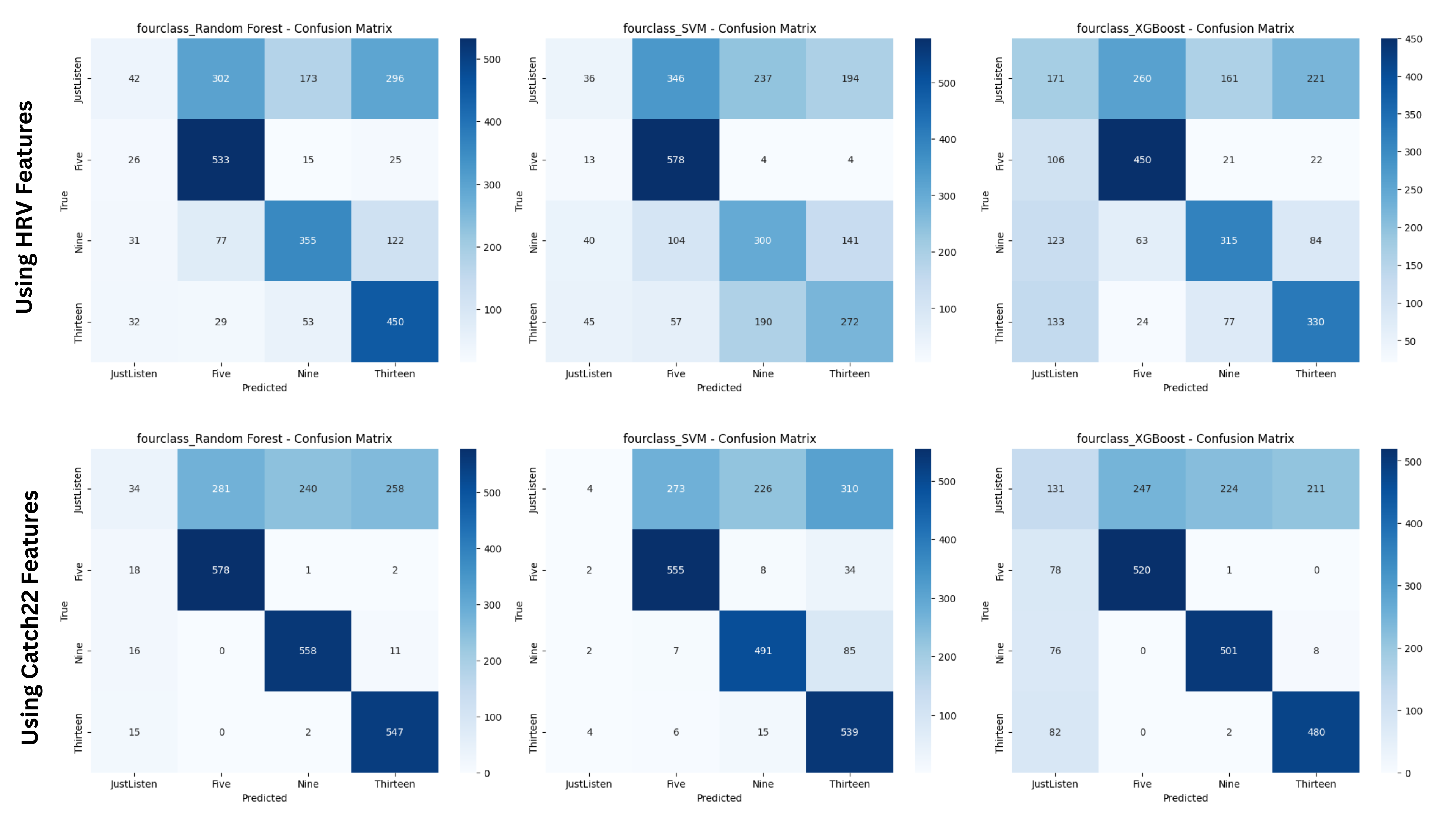} 
    \caption{Confusion Matrices for ECG Four-Class Classifier (FC).}
    \label{fig:C22vsHRV_4class}
\end{figure}

Confusion matrices (Figs. \ref{fig:C22vsHRV_memory},  \ref{fig:C22vsHRV_4class}) show near-perfect MC-classification with Catch22,  while both feature sets misclassify ``Just Listen'' in FC, indicating no strong cardiac baseline signature. Feature-importance (Table \ref{tab:feature_importance}) shows HRV models rely on 1--2 dominant metrics,  whereas Catch22 distributes importance across multiple non-linear features.

\begin{table}[htbp]
\caption{ECG Feature Importance Summary}
\label{tab:feature_importance}
\begin{center}
\resizebox{\columnwidth}{!}{
\begin{tabular}{|c|c|p{3.8cm}|p{3.8cm}|}
\hline
\textbf{Type} & \textbf{Model} & \textbf{Key Features} & \textbf{Interpretation} \\ \hline

HRV & XGBoost &
AutoMutualInfo (f17), AC\_1e\_Crossing (f6), pNN40 (f10) &
Nonlinear cardiac change + short RR-fluctuation sensitivity \\ \hline

HRV & Random Forest &
MeanNN, RMSSD-driven &
Baseline rhythm dominates; less rapid-state sensitivity \\ \hline

Catch22 & XGBoost &
HistogramMode\_5, OutlierPos, ACF\_1/e\_Drop &
State separation from RR distribution + rare peaks + ACF decay \\ \hline

Catch22 & Random Forest &
OutlierPos, LowFreqPower, pNN40 &
Multiple temporal dynamics jointly influence response \\ \hline

Catch22 & XGBoost (re-eval) &
TransitionVar (F18), HistMode\_10, LowFreqPower &
F18 consistently strongest; confirms transition complexity \\ \hline

Fusion & RF (ECG$\rightarrow$EEG) &
HistMode\_10, LowFreqPower, pNN40, TransitionVar &
Stable predictors across neural + cardiac domains \\ \hline
\end{tabular}
}
\end{center}
\end{table}

\subsection{EEG Classification with Classical ML}
EEG-Catch22 classification (Table \ref{tab:eeg_classical}) delivered near-perfect MC accuracy ($\leq$99.82\%). BC/FC performance remained lower,  mirroring ECG behavior and confirming difficulty in baseline separation.

\begin{table}[htbp]
\caption{EEG Classification Performance (Catch22 Features)}
\label{tab:eeg_classical}
\begin{center}
\resizebox{\columnwidth}{!}{
\begin{tabular}{|c|c|c|c|}
\hline
\textbf{Task} & \textbf{Model} & \textbf{Accuracy (\%)} & \textbf{Macro F1} \\ \hline

MC & Random Forest & 99.58 & 1.00 \\ \cline{2-4}
   & SVM           & 98.62 & 0.99 \\ \cline{2-4}
   & XGBoost       & 99.82 & 1.00 \\ \hline

BC & Random Forest & 55.52 & 0.49 \\ \cline{2-4}
   & SVM           & 60.86 & 0.49 \\ \cline{2-4}
   & XGBoost       & 57.54 & 0.50 \\ \hline

FC & Random Forest & 67.68 & 0.61 \\ \cline{2-4}
   & SVM           & 66.74 & 0.60 \\ \cline{2-4}
   & XGBoost       & 63.94 & 0.62 \\ \hline
\end{tabular}
}
\end{center}
\end{table}

\subsection{Deep Learning on Raw Signals}
Raw-signal deep models (Table \ref{tab:deep_learning_classification}) showed ECG-CNN (98.46\%) approaching Catch22-XGBoost (98.68\%),  while EEG - Transformer (99.52\%) matched feature-based performance. DL also failed to resolve BC/FC,  indicating that discriminative patterns lie primarily in cognitive-load epochs rather than baseline separation. BiLSTM underperformed across tasks.

\begin{table}[htbp]
\caption{Deep Learning Classification (Raw Signals)}
\label{tab:deep_learning_classification}
\begin{center}
\resizebox{\columnwidth}{!}{
\begin{tabular}{|c|c|c|c|c|}
\hline
\textbf{Task} & \textbf{Signal} & \textbf{Model} &
\textbf{Accuracy (\%)} & \textbf{Macro F1} \\ \hline

MC & ECG & 1D-CNN      & 98.46 & 0.98 \\ \cline{3-5}
   &     & BiLSTM      & 95.42 & 0.95 \\ \cline{3-5}
   &     & Transformer & 95.71 & 0.96 \\ \cline{2-5}
   & EEG & 1D-CNN      & 99.40 & 0.99 \\ \cline{3-5}
   &     & BiLSTM      & 70.45 & 0.59 \\ \cline{3-5}
   &     & Transformer & 99.52 & 1.00 \\ \hline

BC & ECG & 1D-CNN      & 43.38 & 0.43 \\ \cline{3-5}
   &     & BiLSTM      & 68.25 & 0.41 \\ \cline{3-5}
   &     & Transformer & 31.75 & 0.24 \\ \cline{2-5}
   & EEG & 1D-CNN      & 55.73 & 0.52 \\ \cline{3-5}
   &     & BiLSTM      & 31.58 & 0.24 \\ \cline{3-5}
   &     & Transformer & 68.42 & 0.41 \\ \hline

FC & ECG & 1D-CNN      & 66.69 & 0.61 \\ \cline{3-5}
   &     & BiLSTM      & 66.03 & 0.59 \\ \cline{3-5}
   &     & Transformer & 66.54 & 0.59 \\ \cline{2-5}
   & EEG & 1D-CNN      & 67.52 & 0.61 \\ \cline{3-5}
   &     & BiLSTM      & 65.09 & 0.58 \\ \cline{3-5}
   &     & Transformer & 67.52 & 0.60 \\ \hline
\end{tabular}
}
\end{center}
\end{table}

\subsection{Cross-Modal Fusion Classification}
\begin{figure}[htbp]
    \centering
    \includegraphics[width=0.95\linewidth]{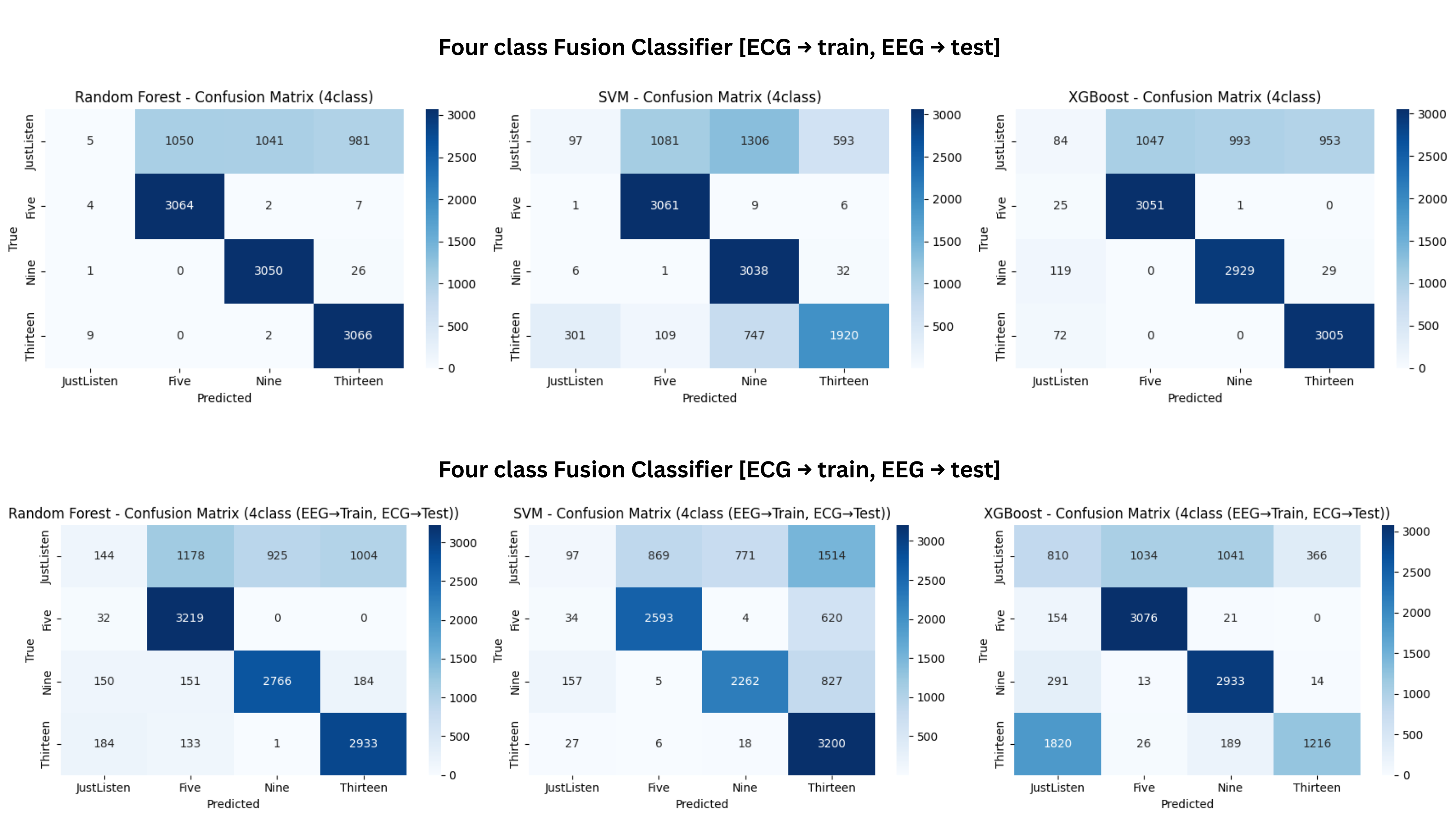} 
    \caption{Confusion Matrices for ECG Four-Class Classifier (FC).}
    \label{fig:Confusion_matrix_Fusion}
\end{figure}

To evaluate cross-modal transferability,  models were trained on Catch22 features from one modality and tested on the other (Table \ref{tab:fusion}). ECG$\rightarrow$EEG transfer during MC achieved 99.62\% accuracy, matching native EEG performance, indicating strong cardiac$\rightarrow$neural generalization. EEG$\rightarrow$ECG transfer also remained high (93.11\%),  confirming bidirectional coupling. BC remained weak,  while FC classification improved under ECG$\rightarrow$EEG (74.63\%),  outperforming native EEG-only models and suggesting ECG features may stabilize neural representations (Fig.~\ref{fig:Confusion_matrix_Fusion}).

\begin{table}[htbp]
\caption{Fusion Classifier (Transfer Learning on Catch22)}
\label{tab:fusion}
\begin{center}
\resizebox{\columnwidth}{!}{
\begin{tabular}{|c|c|c|c|c|}
\hline
\textbf{Task} & \textbf{Transfer Direction} & \textbf{Model} &
\textbf{Accuracy (\%)} & \textbf{Macro F1} \\ \hline

MC & ECG$\rightarrow$EEG & RF      & 99.58 & 1.00 \\ \cline{3-5}
   &                     & SVM     & 90.26 & 0.90 \\ \cline{3-5}
   &                     & XGBoost & 99.62 & 1.00 \\ \cline{2-5}
   & EEG$\rightarrow$ECG & RF      & 93.11 & 0.93 \\ \cline{3-5}
   &                     & SVM     & 82.96 & 0.83 \\ \cline{3-5}
   &                     & XGBoost & 87.67 & 0.87 \\ \hline

BC & ECG$\rightarrow$EEG & RF      & 49.99 & 0.38 \\ \cline{3-5}
   &                     & SVM     & 49.58 & 0.43 \\ \cline{3-5}
   &                     & XGBoost & 50.34 & 0.44 \\ \cline{2-5}
   & EEG$\rightarrow$ECG & RF      & 48.70 & 0.44 \\ \cline{3-5}
   &                     & SVM     & 50.00 & 0.33 \\ \cline{3-5}
   &                     & XGBoost & 49.69 & 0.46 \\ \hline

FC & ECG$\rightarrow$EEG & RF      & 74.63 & 0.64 \\ \cline{3-5}
   &                     & SVM     & 65.94 & 0.58 \\ \cline{3-5}
   &                     & XGBoost & 73.68 & 0.65 \\ \cline{2-5}
   & EEG$\rightarrow$ECG & RF      & 69.69 & 0.62 \\ \cline{3-5}
   &                     & SVM     & 62.69 & 0.56 \\ \cline{3-5}
   &                     & XGBoost & 61.79 & 0.59 \\ \hline
\end{tabular}
}
\end{center}
\end{table}

\section{Discussion}
Our findings provide three core insights. First,  Catch22 features are significantly more informative than standard HRV metrics for ECG-based cognitive load detection,  as they capture nonlinear and dynamical properties rather than relying on variance-based RR statistics.

A primary cause of this is the short-window constraint: all HRV measures like SDNN and RMSSD require $\geq 60$ seconds of RR data for stability, while our 3-second retention window contains too few cardiac cycles (often $\geq 5$) to produce reliable variance estimates. This would explain the poor performance of the HRV measures.

Second,  the primary bottleneck across ECG,  EEG,  ML,  and DL models was not differentiating levels of cognitive load,  but separating cognitive engagement from the passive "Just Listen" condition. Results indicate that the baseline state lacks a strong and consistent physiological signature,  leading to confusion with low-load trials.

Third,  cross-modal transfer learning demonstrated strong heart–brain coupling. The outstanding ECG→EEG projection accuracy confirms that cardiac dynamics can predict neural representations of memory load,  supporting ECG as an accessible surrogate for EEG. EEG→ECG transfer also remained high,  further indicating bidirectional autonomic–cortical interaction. A current limitation is that only auditory-task cognitive load was assessed,  and generalizability to other cognitive domains and real-world noise remains to be validated.

\section{Conclusion}
\label{sec:conclusions}
We investigated whether ECG can serve as a practical substitute for EEG in cognitive load estimation. Using Catch22 features,  classical ML,  deep models,  and bidirectional transfer learning,  we showed that ECG can reliably decode working-memory states while approximating EEG-level accuracy.

The ECG–Catch22–XGBoost classifier achieved 98.68\%,  approaching EEG performance (99.82\%) and outperforming HRV-only features. Cross-modal mapping further revealed that ECG features could predict EEG space with 99.62\% accuracy,  demonstrating strong coupling between cardiac dynamics and neural processing. Despite this,  all models struggled to identify the resting "Just Listen" condition,  indicating absence of a clear physiological baseline signature.

Overall, our work presents ECG as a scalable, interpretable, and wearable alternative for cognitive monitoring that bridges central (EEG) and peripheral (ECG) physiology for real-time load estimation.

\subsection{Future Work}
\label{sec:FutureWork}
Future work will put greater emphasis on better separation of baseline cognition and improved generalization to diverse cognitive demands beyond memory, including attention, fatigue, emotion, and stress states. Hybrid feature representations combining Catch22 with spectral, non-linear, and frequency-domain descriptors will be further explored, together with lightweight CNN-LSTM and transformer-based architectures optimized for real-time wearable deployment via pruning, quantization, or edge-efficient inference.

These strong bidirectional transfer-learning results call for further investigation of heart-brain coupling by means of multimodal neuroimaging techniques such as fMRI/MEG, along with behavioral sources like gaze, speech, and micro-expressions. Combining these with additional wearable-friendly modalities such as PPG, GSR, and respiration may allow richer decoding of states and passive monitoring in a continuous manner. Valuable translation pathways may be extensions towards adaptive cognitive feedback systems, surveillance of workplace stress, and early detection of neuro-cognitive decline.

\section*{Code Availability}
The code and experimental resources used in this study are publicly available at:
\url{https://github.com/AkshaySasi/Unveiling-the-Heart-Brain-Connection-An-Analysis-of-ECG-in-Cognitive-Performance}

\section{Acknowledgment}
The authors express their sincere gratitude to their supervisors and faculty for their continuous guidance, valuable insights, and encouragement throughout this research. We also acknowledge the resources and support received from the University and the availability of the Digit Span dataset through the OpenNeuro platform.
Large Language Models (LLMs), were used only for language refinement, proofreading, and formatting assistance. No part of the experimental design, data analysis, results, or scientific claims was generated by AI. All research ideas, implementations, and interpretations are solely the contribution of the authors.

\end{document}